%% file: acl2023.tex
\pdfoutput=1

\documentclass[11pt]{article}

\usepackage[]{ACL2023}

\usepackage{times}
\usepackage{latexsym}

\usepackage[T1]{fontenc}

\usepackage[utf8]{inputenc}

\usepackage{microtype}

\usepackage{inconsolata}

\definecolor{lightgray}{gray}{0.9}
\colorlet{soulgreen}{green!30}
\definecolor{red}{HTML}{FF0000}
\definecolor{blue}{HTML}{0000FF}
\definecolor{darkgreen}{HTML}{228B22}
\definecolor{dblue}{HTML}{007FFF}
\definecolor{figpurple}{HTML}{D5CEEF }
\definecolor{figblue}{HTML}{C9E3E5}
\definecolor{figred}{HTML}{E3B1B7}

\usepackage{booktabs,arydshln}
\DeclareRobustCommand{\hlpurple}[1]{{\sethlcolor{figpurple}\hl{#1}}}
\DeclareRobustCommand{\hlblue}[1]{{\sethlcolor{figblue}\hl{#1}}}
\DeclareRobustCommand{\hlred}[1]{{\sethlcolor{figred}\hl{#1}}}

\makeatletter
\def\adl@drawiv#1#2#3{%
        \hskip.5\tabcolsep
        \xleaders#3{#2.5\@tempdimb #1{1}#2.5\@tempdimb}%
                #2\z@ plus1fil minus1fil\relax
        \hskip.5\tabcolsep}
\newcommand{\cdashlinelr}[1]{%
  \noalign{\vskip\aboverulesep
           \global\let\@dashdrawstore\adl@draw
           \global\let\adl@draw\adl@drawiv}
  \cdashline{#1}
  \noalign{\global\let\adl@draw\@dashdrawstore
           \vskip\belowrulesep}}
\makeatother

\usepackage{inconsolata}
\usepackage{adjustbox}

\usepackage{blindtext}
\usepackage{graphicx}
\usepackage{capt-of}
\usepackage{booktabs}
\usepackage{varwidth}
\usepackage{multirow}
\usepackage{xspace,mfirstuc,tabulary}

\newcolumntype{L}[1]{>{\raggedright\let\newline\\\arraybackslash\hspace{0pt}}m{#1}}
\newcolumntype{C}[1]{>{\centering\let\newline\\\arraybackslash\hspace{0pt}}m{#1}}

\newsavebox\tmpbox

\newcommand{\dataset}{MEDI\xspace}

\newcommand{\ours}{\textsc{InstructOR}\xspace}

\usepackage{bm}
\usepackage{color}
\usepackage{graphicx}
\usepackage[toc,page]{appendix}
\usepackage{makecell}
\usepackage{boldline}

\usepackage{algorithm}
\usepackage{colortbl}
\usepackage{tabstackengine}
\usepackage{tikz}
\usepackage{relsize}
\usepackage{microtype}
\usepackage{tablefootnote}
\usepackage{silence}
\usepackage{bbm}

\usepackage{tikz}
\usepackage{soul}

\usepackage{fdsymbol}

\definecolor{lightgray}{gray}{0.9}
\colorlet{soulgreen}{green!30}
\definecolor{red}{HTML}{FF0000}
\definecolor{blue}{HTML}{0000FF}
\definecolor{darkgreen}{HTML}{228B22}
\definecolor{dblue}{HTML}{007FFF}
\definecolor{figpurple}{HTML}{D5CEEF }
\definecolor{figblue}{HTML}{C9E3E5}
\definecolor{figred}{HTML}{E3B1B7}

\usepackage{booktabs,arydshln}
\DeclareRobustCommand{\hlpurple}[1]{{\sethlcolor{figpurple}\hl{#1}}}
\DeclareRobustCommand{\hlblue}[1]{{\sethlcolor{figblue}\hl{#1}}}
\DeclareRobustCommand{\hlred}[1]{{\sethlcolor{figred}\hl{#1}}}

\usepackage[noend]{algpseudocode}
\algnewcommand{\parState}[1]{\State%
    \parbox[t]{\dimexpr\linewidth-\algmargin}{\strut\hangindent=\algorithmicindent \hangafter=1 #1\strut}}

\algrenewcommand\algorithmicindent{1.0em}%

\newcommand{\com}[1]{}

\definecolor{magenta}{HTML}{F3DFF1}
\definecolor{red}{HTML}{FF0000}
\definecolor{hlgreen}{HTML}{D5E8D4}
\definecolor{figblue}{HTML}{DAE8FC}

\usepackage{algorithm}
\usepackage{colortbl}
\usepackage{tabstackengine}
\usepackage{tikz}
\usepackage{relsize}
\usepackage{microtype}
\definecolor{magenta}{HTML}{F3DFF1}
\definecolor{hlgreen}{HTML}{ccfcc4}
\definecolor{figblue}{HTML}{e7f2fe}

\usepackage{pifont}

\usepackage{listings}
\definecolor{mymauve}{rgb}{0.58,0,0.82}
\usepackage{paralist, tabularx}

\title{One Embedder, Any Task: Instruction-Finetuned Text Embeddings}

\usepackage{fdsymbol}

\author{
\textbf{Hongjin Su}$^\spadesuit$\thanks{\ \ Equal contribution.} \ \ 
        \textbf{Weijia Shi}$^{\clubsuit}$\footnotemark[1] \ \ 
        \textbf{Jungo Kasai}$^\clubsuit$ \ \ 
        \textbf{Yizhong Wang}$^{\clubsuit}$ \ \ 
        \textbf{Yushi Hu}$^{\clubsuit}$\\ 
         \textbf{Mari Ostendorf}$^{\clubsuit}$
        \ \ 
         \textbf{Wen-tau Yih}$^{\diamondsuit}$
        \ \ 
        \textbf{Noah A.\ Smith}$^{\clubsuit\heartsuit}$\ \ 
        \textbf{Luke Zettlemoyer}$^{\clubsuit\diamondsuit}$\ \ 
        \textbf{Tao Yu}$^{\spadesuit}$
       \\ 
  $^\spadesuit$The University of Hong Kong 
  \quad
  $^\clubsuit$University of Washington
  \quad
  $^\diamondsuit$Meta AI
  \\
  $^\heartsuit$Allen Institute for AI
  \\
  {\tt \{hjsu,tyu\}@cs.hku.hk,}\ \ 
  {\tt \{yushihu,ostendor\}@uw.edu} \ \ 
  {\tt scottyih@meta.com}
 \\
  {\tt \{swj0419,jkasai,yizhongw,nasmith,lsz\}@cs.washington.edu}
}

\begin{document}
\maketitle
\input{text/abstract.tex}
\input{text/intro.tex}
\input{text/method2.tex}

\input{text/experiments}

\input{text/analysis}
\input{text/related}
\input{text/future}
\input{text/limitations}
\input{text/ack}

\nocite{Ando2005,augenstein-etal-2016-stance,andrew2007scalable,rasooli-tetrault-2015,goodman-etal-2016-noise,harper-2014-learning}

\bibliography{custom}
\bibliographystyle{acl_natbib}

\appendix

\input{text/appendix}

\end{document}

%% file: text/abstract.tex
\begin{abstract} 
We introduce \ours, a new method for computing text embeddings given task instructions: every text input 
is embedded together with instructions explaining the use case (e.g., task and domain descriptions).
Unlike encoders from prior work that are more specialized, \ours is a single embedder that can generate text embeddings tailored to different downstream tasks and domains, \emph{without} any further training.
We first annotate instructions for 330 diverse tasks and train \ours on this multitask mixture with a contrastive loss.
We evaluate \ours on 70 embedding evaluation tasks (66 of which are \emph{unseen} during training), ranging from classification and information retrieval to semantic textual similarity and text generation evaluation.
\ours, while having an order of magnitude fewer parameters than the previous best model, achieves state-of-the-art performance, with an average improvement of 3.4\% compared to the previous best results on the 70 diverse datasets.
Our analysis suggests that \ours is robust to changes in instructions, and that instruction finetuning mitigates the challenge of training a single model on diverse datasets.
Our model, code, and data are available at \url{https://instructor-embedding.github.io}.
\end{abstract}

%% file: text/intro.tex
\section{Introduction}
\begin{figure}[h]
    \centering
    \includegraphics[scale=0.47]{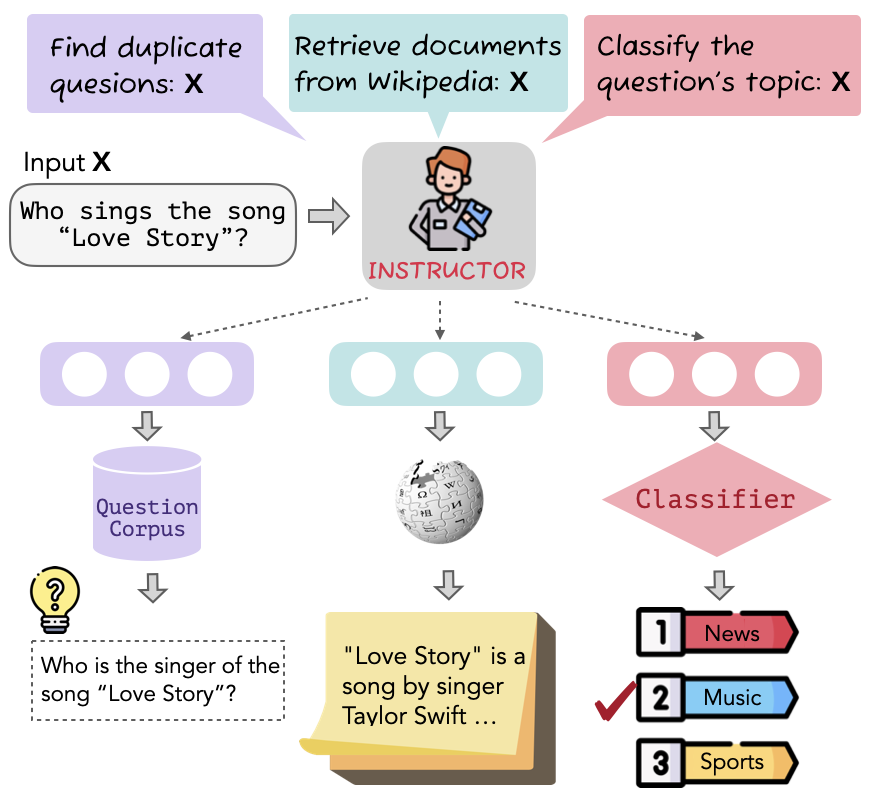} 
    \caption{
    At execution time, \ours generates embeddings based on both the text input and the task instruction. The same input (e.g., \textit{Who sings the song ``Love Story''?}) will be encoded into different embeddings, depending on the end task (e.g., \hlpurple{duplicate question detection}, \hlblue{information retrieval}, and \hlred{topic classification}). 
    } 
    \label{fig:instructor}
\end{figure}

\begin{figure*}[h]
    \centering
   \includegraphics[width=0.9\linewidth]{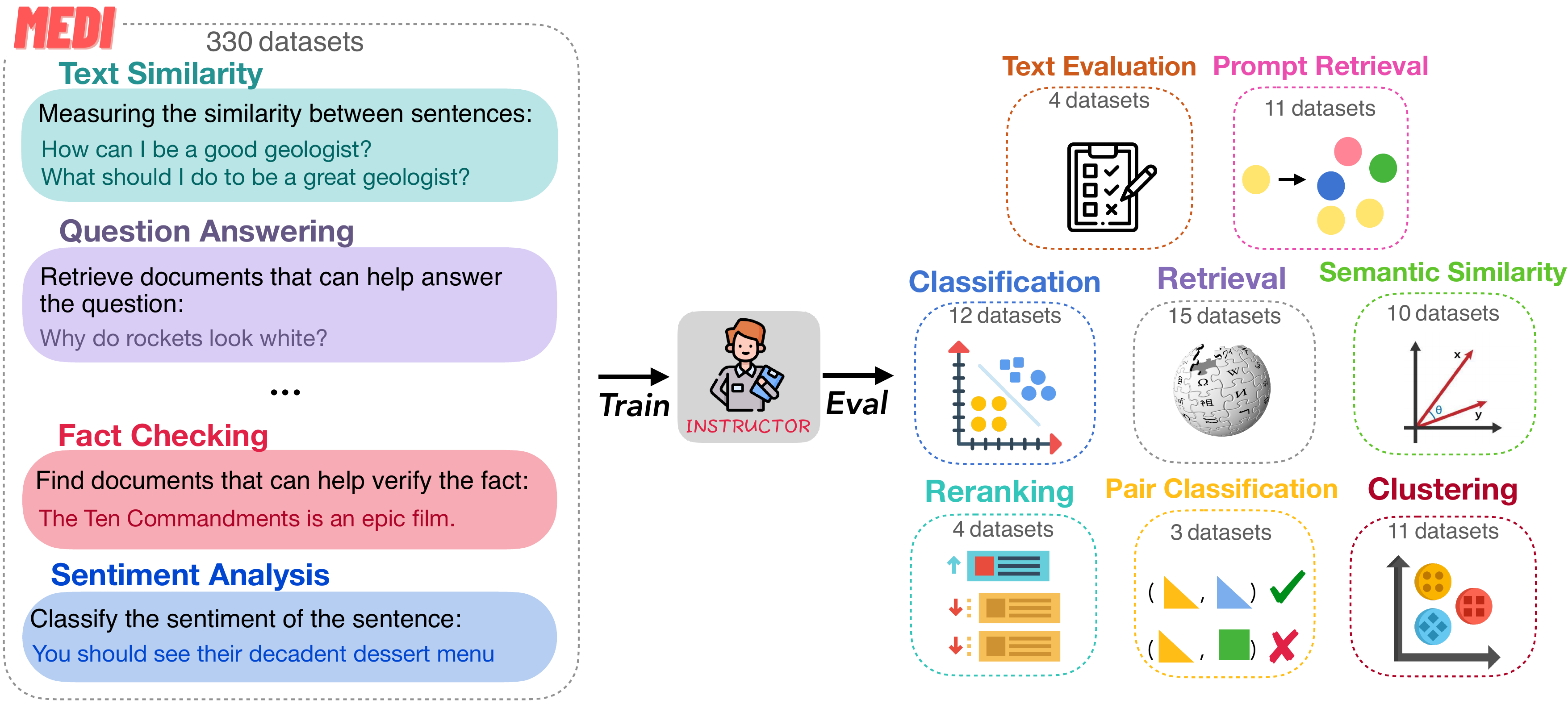}
   \caption[Caption for LOF]
    {\ours training and evaluation pipeline. 
    \ours is a single embedding model that takes not only text inputs but also task instructions, thereby creating task- and domain-aware embeddings.
    It is trained on a multitask mixture of 330 diverse datasets with human-written task instructions (\dataset dataset, \S\ref{section:dataset}).
    After training on \dataset (left), \ours is evaluated on a variety of 70 embedding datasets (66 of which are not seen during training), spanning various downstream applications (right).
    \ours outperforms the prior best model by an average of 3.4\% over the 70 diverse datasets.
    }
    \label{fig:pipeline}
\end{figure*}

Text embeddings represent discrete text inputs (e.g., sentences, documents, and code) as fixed-sized vectors that can be used in many downstream tasks.
These tasks include semantic textual similarity~\cite{semeval12,sick14,cer2017semeval,lin:2018:semantic}, information retrieval \cite{Mitra0C17,karpukhin-etal-2020-dense,izacard2022unsupervised}, automatic text evaluation \cite{bertscore,sellam2020bleurt,clipscore}, prompt retrieval for in-context learning \cite{liu-etal-2022-makes,rubin2022,Selective_Annotation}, and beyond.
Recently, we have seen dramatic advances in learning text embeddings~\cite{Kiros15,conneau-etal-2017-supervised,logeswaran2018an,reimers-gurevych-2019-sentence,gao-etal-2021-simcse,gtr21,ni2022sentence} that perform well on their intended tasks or datasets.

However, most existing embeddings can have significantly degraded performance when applied to new tasks or domains~\cite{thakur2021beir,mteb}.
For example, DPR~\cite{karpukhin-etal-2020-dense} is stronger for retrieval than text similarity tasks, and vice versa for SimCSE \cite{gao-etal-2021-simcse}.
Moreover, existing embeddings usually perform poorly when applied to the same type of task but in different domains such as medicine and finance.
A common method to address this issue is to further finetune the embeddings on datasets in downstream tasks and domains, which often requires a lot of annotated data \cite{gururangan-etal-2020-dont}.
In this paper, we hypothesize that text embeddings (even for the \emph{same} text input) can be adjusted to different downstream applications using task and domain descriptions, \emph{without} further task- or domain-specific finetuning.

We introduce \ours (\textbf{Instruct}ion-based \textbf{O}mnifarious \textbf{R}epresentations), a single multitask model that generates task- and domain-aware embeddings given a text input and its task instructions. 
It achieves state-of-the-art performance on 
massively many downstream embedding tasks without any training.
At the core of our approach is instruction-based finetuning~\cite{Zhong2021AdaptingLM,min-etal-2022-metaicl,Sanh2022Multitask,Wei2022FinetunedLM}: we embed every input together with its end task and domain instruction, departing from prior approaches to embeddings that only take text input.
\ours embeds the same input into different vectors for different end goals (e.g., \textit{Who sings the song ``Love Story''?} is embedded into three different vectors for different tasks in Fig.\ \ref{fig:instructor}).
As shown in Fig.~\ref{fig:pipeline}, \ours is trained on \dataset,  
our new collection of 330 text embedding datasets newly annotated with human-written task instructions (\S\ref{section:dataset}). 
We train \ours with a contrastive loss over all datasets that maximizes the similarity between semantically related text pairs while minimizing unrelated pairs.

We extensively evaluate \ours on diverse domains (e.g., finance, medicine, and news) and a variety of downstream applications (a total of 70 embedding evaluation datasets, including 66 \emph{not} seen during training), spanning classification, semantic textual similarity, information retrieval, text generation evaluation, and prompt retrieval for in-context learning.
\ours significantly outperforms prior state-of-the-art embedding models by an average of 3.4\% over the 70 diverse datasets.
\ours also outperforms a variant that is trained \emph{without} task instructions (\S\ref{section:analysis}), demonstrating the importance of instructions to create task-aware embeddings.
Our analysis shows that instruction finetuning addresses the challenge of training a \emph{single} model on \emph{diverse} datasets (\S\ref{section:ablation_datasets}).
Further, we demonstrate 
that the task diversity of \dataset makes the performance of \ours particularly robust to paraphrases in instructions (\S\ref{section:instruction-robustness}).
Overall, these results strongly suggest that instruction finetuning should be adopted broadly for text embeddings, which we support by sharing all of our models and code. 

%% file: text/method2.tex
\section{\ours}

\label{section:instruction-base_embeddings}
\ours encodes inputs together with task instructions, thereby providing task-specific representations that can be used for many downstream language tasks, \emph{without} any additional training.
Here we introduce the architecture of \ours (\S\ref{section:architecture}), present how we perform multitask instruction-based finetuning (\S\ref{section:training_objective}), and describe how we collect and annotate the \dataset training data (\S\ref{section:dataset}). By default, we refer "task" to a dataset, and use them interchangeably throughout the paper, while a "task category", such as Retrieval, includes many tasks.

\subsection{Embedding Architecture}
\label{section:architecture}
We build \ours, based on the single encoder architecture~\cite{izacard2021leveraging, gtr21, ni2022sentence}. Following prior work \cite{gtr21, ni2022sentence}, we use GTR models as the backbone encoder (GTR-Base for \ours-Base, 
GTR-Large for \ours, GTR-XL for {\ours}-XL).
The GTR models are initialized from T5 models, pretrained on a web corpus, and finetuned on information search datasets.
The availability of different sizes in the GTR model family allows us to explore the scaling behaviors of instruction-finetuned embedding models.
Given an input text $x$ and a task instruction $I_x$, \ours encodes their concatenation $I_x \oplus x$.
We then generate a fixed-sized, task-specific embedding $\textbf{E}_I(I_x , x)$ by applying mean pooling to the last hidden representations over the tokens in $x$. 

\subsection{Training Objective}
\label{section:training_objective}

\ours is trained by formulating a wide variety of tasks as a text-to-text problem of distinguishing good/bad candidate outputs $y \in \{y^+ , y^-_i \}$ given an input $x$, where
a training sample corresponds to the tuple $(x, I_x, y, I_y)$, with $I_x$ and $I_y$ being instructions associated with $x$ and $y$, respectively. 
For example, in a retrieval task, $x$ is a query, and good/bad $y$ is a relevant/irrelevant document from some document collection.
For a textual similarity task, the input and output have a similar form and typically come from the same source collection. 
For a classification task, training samples can be formed by choosing $y$ as text sequences associated with the same vs.\ different classes for good vs.\ bad examples (Details about pair construction are in \S\ref{section:dataset}).
The input and output instructions depend on the task. For {\bf symmetric} tasks such as textual similarity, where the input and output have the same form and encoding objective, the instructions are the same. For {\bf asymmetric} tasks such as retrieval, where the input is a single sentence query and the output is a document, the instructions reflect that difference.

The goodness of candidate $y$ for input $x$ is given by similarity $s(x, y)$ that is the cosine between their \ours embeddings: 
\begin{align*}
s(x, y) = \cos (\textbf{E}_I(I_x \oplus x), \textbf{E}_I(I_{y} \oplus y))
\end{align*}
Following \citet{gtr21}, we maximize the similarity between positive pairs $(x, y^+)$ and minimize negative pairs $\{(x, y^-_{i})\}_{i=1}^k$, where $k$ denotes the number of negative pairs per positive pair. 
Specifically, our training objective is:
\begin{align*}
\mathcal{L} = \frac{e^{s(x, y^+)/\gamma}}{\sum_{y \in \mathcal{B}} e^{s(x, y)/\gamma}},
\end{align*}
where $\gamma$ is the softmax temperature and $\mathcal{B}$ is a union of $(x, y^{+})$ and $\{(x, y^{-}_i)\}_{i=1}^k$.
Further following~\citet{gtr21}, we compute the same loss with $x$ and $y$ swapped and add it to the previous loss (i.e., bidirectional in-batch sampled loss).

\input{table/instruction.tex}

\subsection{\dataset: Multitask Embedding Data with Instructions} \label{section:dataset}
There are no existing datasets that consist of a variety of tasks for embedding training with instructions.
We thus construct a collection of \textbf{330} datasets with instructions across diverse task categories and domains: \textbf{M}ultitask \textbf{E}mbeddings \textbf{D}ata with \textbf{I}nstructions (\dataset).

\input{table/main.tex}
\paragraph{Data Construction}
We build MEDI by combining 300 datasets from Super-NaturalInstructions~(super-NI; \citealp{natural_instructions_v2}) with
30 datasets from existing collections designed for embedding training. 

The super-NI datasets come with natural language instructions, but positive and negative pairs are not provided.
We construct these pairs by using Sentence-T5 embeddings \cite{ni2022sentence},\footnote{We do not include instruction for Sentence-T5 as it is not fine-tuned with instructions.} denoted with $\textbf{E}(\cdot)$.
For the classification datasets, we calculate the pairwise cosine similarity between examples based on input text embeddings $\cos(\textbf{E}(x_i),\textbf{E}(x_j))$.
An example $x_i$ with a high similarity to $x_j$ is used to create a positive pair if both examples have the same class label ($y_j^+=y_i$), and a negative pair if the labels differ ($y_j^- \neq y_i$).  For the remaining tasks where the output labels are text sequences, the following scores are first computed:
\begin{align*}
s_{pos} = \cos(\textbf{E}(x_i),\textbf{E}(x_j))+\cos(\textbf{E}(y_i),\textbf{E}(y_j))
\end{align*}
\begin{align*}
s_{neg} = \cos(\textbf{E}(x_i),\textbf{E}(x_j))-\cos(\textbf{E}(y_i),\textbf{E}(y_j))
\end{align*}
We select example pairs with the highest $s_{pos}$ as positive pairs and highest $s_{neg}$ as hard negative pairs.
We use one hard negative together with in-batch sampled negatives in the training.
Our later analysis shows that the training data from super-NI particularly improve the instruction robustness in evaluation due to the diverse task definitions (\S\ref{section:instruction-robustness}).

The other 30 embedding training datasets come from the Sentence Transformers embedding data,\footnote{\url{https://huggingface.co/datasets/sentence-transformers/embedding-training-data}.} KILT~\cite{petroni-etal-2021-kilt}, and MedMCQA~\cite{pmlr-v174-pal22a}.
These 30 datasets already contain positive pairs; a few of them, such as MSMARCO~\cite{bajaj2016ms} and Natural Questions \cite{kwiatkowski2019natural}, also contain hard negative pairs.
Following \citet{gtr21}, we use four negative pairs (hard or in-batch negatives) during the model finetuning process.
Since all of these datasets do not have instructions, we develop a unified instruction template and manually write a specific prompt for each dataset, as described next.\footnote{All prompts are reviewed by multiple authors independently to make sure they consistently follow our template.}
We release these instructions together with our \dataset data.

\paragraph{Instruction Annotation}
\label{section:unified_instruction_format}
Each training instance from \dataset is a tuple $(x, I_x, y, I_y)$, where
%
the natural language instructions $I_x$ and $I_y$ describe how the embeddings of $x$ and $y$ are used for the task. 
For example, in open-domain QA (e.g., Natural Questions in Table~\ref{tab:instruct}), $I_x$ is ``Represent the Wikipedia question for retrieving supporting documents; Input: ,'' and $I_y$ is ``Represent
the Wikipedia document for retrieval; Input: .''

To make instructions consistent across all datasets in \dataset, we design a unified instruction format that consists of the following parts (see Table \ref{tab:instruction_examples} in the appendix for instances of each part): 
\begin{compactitem}
    \item \textbf{Text Type} specifies the type of input text that we encode using the embedding model.
    For example, for an open-domain QA task, the input type of the query is a question, while the input type of the target is a document.  
    \item \textbf{Task Objective (Optional)} describes the objective of how the input text is used in a task. 
    For example, for a classification task, the task objective is to classify the sentence into some category, while the task objective of the retrieval is to retrieve a relevant document. 
    Because not all sentences are associated with a specific task (e.g., STS targets general encoding), we make this part optional.
    \item \textbf{Domain (Optional)} describes the task domain. For example, for NewsIR, the domain of the task is news. 
    Because not all tasks specify a domain (e.g., STS deals with general statements),this part is also optional. 
\end{compactitem}
The final instruction takes the following format: 
\noindent ``\textsc{Represent the (\textbf{Domain}) \textbf{Text Type} for \textbf{Task Objective}:}." Appendix~\ref{tab:full_instructions2} shows instructions for each dataset in \dataset.



%% file: table/instruction.tex
\begin{table*}[h]
\small
\centering
\begin{tabularx}{\linewidth}{m{1.3cm}ccm{8cm}}
\toprule[.1em]
\textbf{Task type} & \# of Datasets & Task & Instruction   \\
\midrule[.1em]
Retrieval& 15 & Natural Question (BEIR) & \textbf{\textit{Query instruction:}} Represent the Wikipedia question for retrieving supporting documents:, \textbf{\textit{Doc instruction:}} Represent the Wikipedia document for retrieval:\\
\midrule[.1em]
Reranking& 4 & MindSmallReranking & \textbf{\textit{Query instruction:}} Represent the News query for retrieving articles: \textbf{\textit{Doc instruction:}} Represent the News article for retrieval: \\
\midrule[.1em]
Clustering&11 & MedrxivClusteringS2S & Represent the Medicine statement for retrieval:  \\
\midrule[.1em]
Pair Classification & 3 & TwitterSemEval2015 & Represent the Tweet post for retrieving duplicate comments: \\
\midrule[.1em]
Classification& 12 & ImdbClassification & Represent the Review sentence for classifying emotion as positive or negative: \\
\midrule[.1em]
STS& 10 & STS12 & Represent the statement: \\
\midrule[.1em]
Summarization & 1 & SummEval & Represent the Biomedical summary for retrieving duplicate summaries: \\
\midrule[.1em]
Text Evaluation & 3 & Mscoco & Represent the caption for retrieving duplicate captions: \\
\midrule[.1em]
Prompt Retrieval & 11 & GeoQuery & Represent the Geography example for retrieving duplicate examples: \\
\bottomrule[.1em]
\end{tabularx}
\caption{
Instruction examples for evaluation datasets.
Our embedding evaluation includes 70 diverse datasets in 9 different downstream applications, ranging from classification and semantic textual similarity to information retrieval and text generation evaluation. 
The first two tasks are \textbf{asymmetric} and require two distinct instructions.
Instructions for the \dataset training data can be found in Tables \ref{tab:full_instructions1} and \ref{tab:full_instructions2} in the appendix.
}
\label{tab:instruct}
\end{table*}

%% file: table/main.tex
\begin{table*}[h]
\small
 \addtolength{\tabcolsep}{-2.0pt} 
\centering
\begin{tabular}{lc c ccccccc   c  | c    }
\toprule
  \textbf{\textit{Benchmark}} &   \multicolumn{8}{c}{\textbf{MTEB}}&  \textbf{Billboard}  &  \textbf{Prompt} & \textbf{Avg.} \\
 \cmidrule(lr){2-9}
 \cmidrule(lr){10-10}
 \cmidrule(lr){11-11}
 \cmidrule(lr){12-12}
  \textbf{\textit{Task category}} &  Retri. & Rerank & Cluster & Pair. & Class. & STS & Sum. & Avg.   & Text Eval.  & Retri. &  \\
\textbf{  \textit{\# datasets}}  &  15 &  4 & 11 & 3  & 12 & 10 & 1 &  56  & 3 & 11 & 70 \\
\midrule
\multicolumn{2}{l}{\textbf{Small Models for reference ($<$500M)}} & &  &  & &&& &&  & \\
SimCSE (110M) & 21.9 & 47.5 &  33.4 & 73.7 & 67.3 & 79.1 & 23.3 & 48.7 & 29.4 & 58.3 & 48.2 \\
coCondenser (110M) & 33.0 & 51.8  & 37.6 & 81.7 & 64.7 & 76.5 & 29.5 & 52.4 & 31.5 & 59.6 & 51.8 \\
Contriever (110M) & 41.9 & 53.1 & 41.1 & 82.5 & 66.7  & 76.5 & 30.4 & 56.0 & 29.0 & 57.3 & 53.2 \\
\midrule
GTR-Large (335M) & 47.4 & 55.4 & 41.6 & 85.3 & 67.1  & 78.2 & 29.5 & 58.3 & 31.2 & 59.8 & 55.1 \\
\ours(335M) & \textbf{47.6} & \textbf{57.5} & \textbf{45.3} & \textbf{85.9} & \textbf{73.9} & \textbf{83.2} & \textbf{31.8} & \textbf{61.6} & \textbf{36.9} &  \textbf{63.2} & \textbf{58.4}\\
Relative gain (\%) & \textcolor{blue}{+0.4} & \textcolor{blue}{+4.5} & \textcolor{blue}{+8.9} & \textcolor{blue}{+0.7} & \textcolor{blue}{+10.1} & \textcolor{blue}{+6.4} & \textcolor{blue}{+7.8} & \textcolor{blue}{+5.7}  & \textcolor{blue}{+18.3}  & \textcolor{blue}{+5.7} & \textcolor{blue}{+5.9}\\
\midrule[.02em]
\midrule[.02em]
\multicolumn{2}{l}{\textbf{Large Models for reference($\geq$500M)}} & &  &  & &&& &&  &\\
Sent-T5-XXL (4.8B) & 42.2 &  56.4 & 43.7 & 85.1 & \textbf{73.4} & 82.6 & 30.1 & 59.5 & 33.9 & 61.5 & 56.5 \\
GTR-XXL (4.8B) & 48.1 &  56.7 & 42.4 & 86.1  & 67.4 & 78.4 & 30.6& 58.9 & 32.0 & 60.8 & 55.8\\
SGPT-NLI (5.8B) & 32.3 & 52.3  & 37.0 & 77.0 & 70.1 & 80.5 & 30.4 & 53.7 & 29.6 & 57.9 & 51.9 \\

\midrule
GTR-XL (1.5B) & 48.0 & 56.0 & 41.5 & 86.1 & 67.1  & 77.8 & 30.2 & 58.4 & 32.0 & 60.4 & 55.5 \\
\ours-XL (1.5B) & \textbf{49.3} & \textbf{57.3} & \textbf{44.7} & \textbf{86.6} & 73.2 & \textbf{83.1} & \textbf{32.0} & \textbf{61.8} & \textbf{34.1} &  \textbf{68.6} & \textbf{58.8}\\
Relative gain (\%) & \textcolor{blue}{+2.7} & \textcolor{blue}{+2.3} & \textcolor{blue}{+7.7} & \textcolor{blue}{+0.6} & \textcolor{blue}{+9.1} & \textcolor{blue}{+6.9} & \textcolor{blue}{+6.0} & \textcolor{blue}{+5.8}  & \textcolor{blue}{+6.6}  & \textcolor{blue}{+13.6} & \textcolor{blue}{+5.9}\\

\bottomrule
\end{tabular}
\caption{
Results on the massive text embedding benchmark (MTEB; \citealp{mteb}), Billboard \cite{billboard}, and prompt retrieval \cite{Selective_Annotation}.
The last column averages performance scores over 9 categories (7 from MTEB, 1 from Billboard, and 1 from prompt retrieval).
Out of the 70 evaluation datasets, 66 (50 from MTEB, 3 from BillBoard, and 11 from prompt retrieval) are unseen tasks during finetuning.
Retri., Pair., Class., Sum., Text Eval. refer to retrieval, pair classification, classification, summarization, and text evaluation, respectively. 
Instruction finetuning improves performance by 5.9\% compared to GTR(335M/1.5B) and achieves 3.4\% and 4.1\% performance gains over the state-of-the-art model Sent-T5-XXL for INSTRUCTOR (335M/1.5B). The relative gain (\%) indicates \ours's improvement relative to the original GTR model of the same size.
}
\label{tab:mteb_results}
\end{table*}

%% file: text/experiments.tex
\section{Experiments}
\label{section:experiments}
We train \ours on the MEDI data and evaluate it on a wide range of 70 downstream tasks.
Specifically, we use the MTEB benchmark from recent work \cite{mteb}, which consists of 56 datasets over 7 diverse task categories, such as classification, reranking, and information retrieval. 
We then further apply \ours to prompt retrieval for in-context learning and text generation evaluation.
In all three settings, \ours achieves the state-of-the-art performance. See Appendix \S\ref{sec:training_settings} and \S\ref{sec:eval_settings} for our detailed settings.



\subsection{Main Results}
\label{section:results}
Table \ref{tab:mteb_results} presents the results from \ours and the baselines over the three benchmarks: MTEB, Billboard, and prompt retrieval.
We conduct head-to-head comparison between \ours and GTR models with the same size.
We also include the performance of other representative models for reference, while they are not meant for direct comparison.

\ours achieves the best performance on all three benchmarks on average.
Compared to GTR-Large (335M), from which \ours is initialized, instruction finetuning enhances the performance by 5.7\%, 18.3\%, and 5.7\% in MTEB, Billboard, and prompt retrieval respectively. 
Specifically, among all task categories, \ours (335M) demonstrates large improvements over GTR-Large on the text evaluation (18.3\%), classification (10.1\%), and clustering tasks (8.9\%).
Particularly noteworthy is \ours's performance compared to the previous state-of-the-art model, Sent-T5-XXL (58.4 vs.\ 56.5 on average), despite the fact that \ours has one order of magnitude fewer parameters (335M vs.\ 4.8B).

As expected, the retrieval-based models (e.g., GTR-XXL) show strong performance on retrieval and reranking but significantly lag behind on STS and classification.
Conversely, similarity-based models (e.g., Sent-T5-XXL) perform well on STS, classification, and text evaluation, but not on retrieval.
It suggests that these baselines tend to generate specialized embeddings that only excel at certain tasks, while \ours provides universal embeddings that perform well on diverse task categories. 

%
%
%
%
%
%
%
%
%
%
%

%% file: text/analysis.tex
\section{Analysis and Ablations}
We demonstrate \ours enables universal text embeddings for many diverse tasks.
Here we analyze our results from various perspectives: the importance of instructions (\S\ref{section:ablation_datasets}), instruction robustness (\S\ref{section:instruction-robustness}) and complexity (\S\ref{section:complexity}), model sizes (\S\ref{section:model_size}), domain shifts (\S\ref{section:domain_shift}), and qualitative analysis (\S\ref{section:qual_analysis}).
By default, we report average performance across all categories.

\begin{figure}[!h]
    \centering
    \includegraphics[scale=0.6]{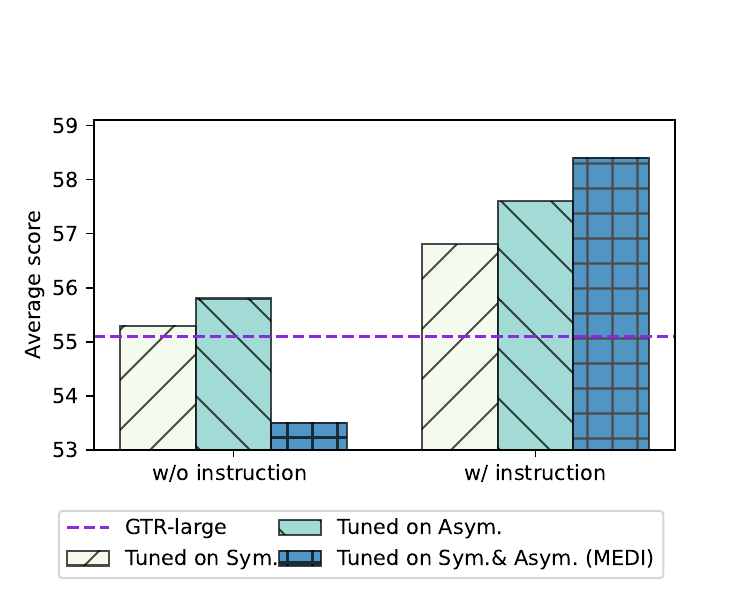} 
    \caption{
    Average (by category) performance of \ours (with and without instructions) over three types of training data: symmetric data, asymmetric data, or both (entire MEDI). The model finetuned with instructions on both data is the original \ours model.
    The diverse training data with both types \textbf{hurt} the performance when finetuned without instructions but \textbf{improve} when instructions are used. 
    }
    \label{fig:instruction}
\end{figure}
\label{section:analysis}

\subsection{Instructions Enable Diverse Training}
\label{section:ablation_datasets}
Here we analyze the importance of instructions when training data are diverse.
We first split \dataset into symmetric (e.g., text similarity) and asymmetric groups (e.g., open-domain QA), as defined in \S\ref{section:unified_instruction_format} (see Table \S\ref{tab:data_statistics} in the appendix for details about the symmetric and asymmetric groups). We then train \ours \emph{with} or \emph{without} instructions on each group separately.

 As shown in Fig.\ \ref{fig:instruction},  \ours finetuned \emph{without} instructions yields performance similar to or better than the original GTR model (dotted line), if the data are symmetric or asymmetric \emph{only}. However, \ours suffers if finetuned without task instructions on the combination of both types of data (entire \dataset).
In contrast, finetuning with instructions enables the model to benefit from the combination of symmetric and asymmetric data (see that the rightmost bar gets additive performance gains from the asymmetric and symmetric tasks).
This result demonstrates the importance of instruction finetuning when diverse data are used for embedding training.
Note that training on symmetric tasks only without instructions is similar to Sent-T5.   
Similarly, training on asymmetric tasks only without instructions is similar to GTR, which is also trained on asymmetric open-domain QA datasets.
Departing from these prior methods, instruction-based finetuning enables diverse training on both types.

\subsection{Instruction Robustness}
\label{section:instruction-robustness}

\begin{figure}[!h]
    \centering
    \includegraphics[scale=0.84]{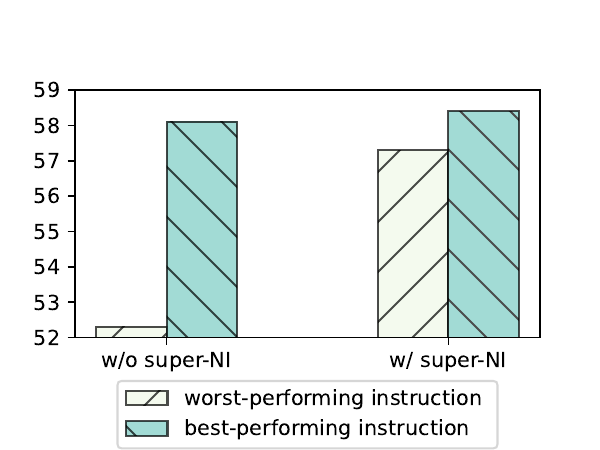} 
    \caption{
    Comparison of the model performance across five paraphrased instructions. W/o super-NI (w/ super-NI) refers to the inclusion (exclusion) of the 300 datasets from Super-NaturalInstructions in \dataset.
    These diverse datasets with task instructions improve the robustness of \ours to instruction paraphrases (i.e., smaller performance gaps between best- and worst-performing instructions).
    }
    \label{fig:prompt_robustness}
\end{figure}

Previous work~\cite{Sanh2022Multitask,zhou2022prompt} shows that instruction-finetuned language models are not robust to paraphrased instructions.
Here we measure \ours's robustness to variation in human-written instructions. 

Specifically, we write five paraphrased instructions 
for all evaluation datasets (Table \ref{tab:prompt_variants} in Appendix) and measure \ours's performance gap between the best-performing and the worst-performing instructions.
Fig.\ \ref{fig:prompt_robustness} shows that 
inclusion of 300 super-NI datasets is critical to the robustness of {\ours}.
Removing these datasets from training (w/o super-NI) substantially increases the performance gap between the best- and worst-performing instructions, suggesting that super-NI's diverse instructions help the model handle different formats and styles. 


\subsection{Complexity of Instructions}
\label{section:complexity}

\begin{figure}[!h]
    \centering
    \includegraphics[scale=0.32]{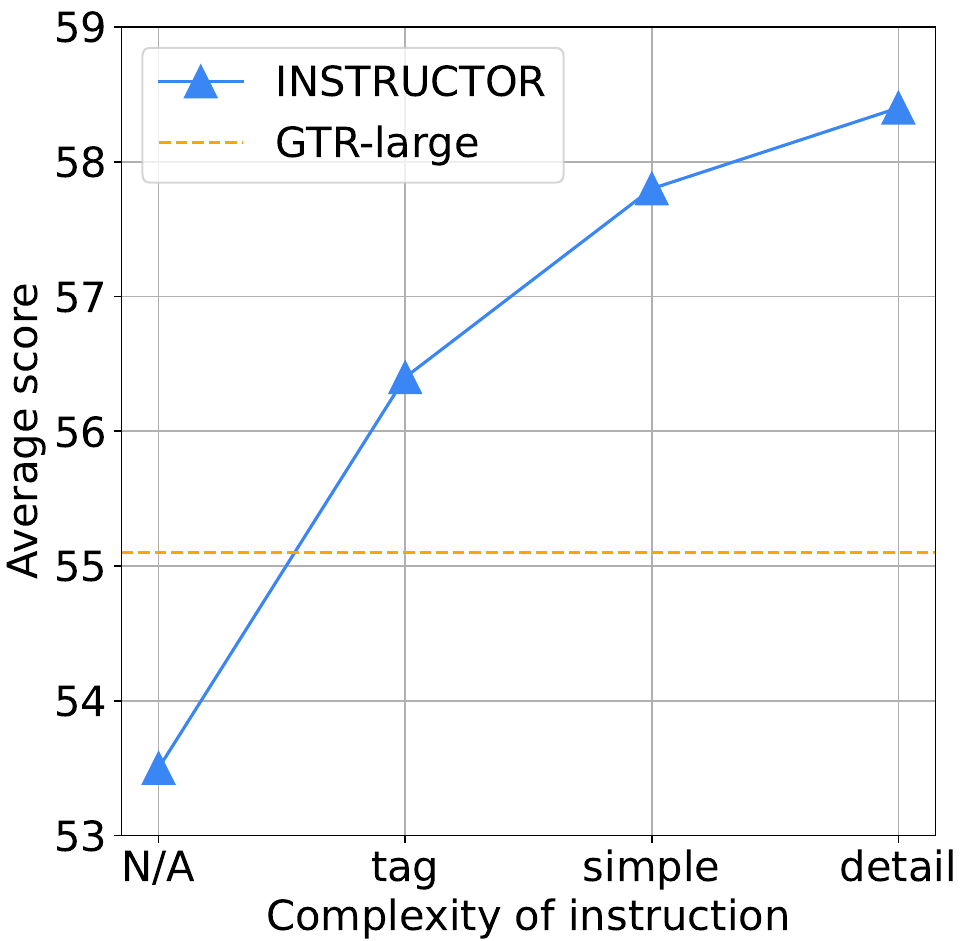} 
    \caption{
    Average performance over varying degrees of instruction details.
    As the instructions become more detailed, the performance improves.
    N/A: no instructions are given; tag: dataset names are prepended; simple: one word or two for the task domain are given (e.g., Wikipedia); and detailed: our proposed instructions (\S\ref{section:unified_instruction_format}).
    %
    }
    \label{fig:instruction_detail}
\end{figure} 

Here we further analyze the role of instructions over varying degrees of their complexity.
Specifically, we consider four levels of instruction complexity: N/A (no instructions), dataset tags, simple instructions, and detailed instructions (the original instruction format, \S\ref{section:unified_instruction_format}).
In the dataset tag setup, each example is prepended with its dataset name. 
For instance, on the Natural Questions dataset, the query is formatted as \textit{"Natural Questions; Input: who sings the song Love Story"}).
In the simple instruction setup, we use one or two words to describe the domain (e.g., for Natural Questions, the input query is \textit{Wikipedia Questions; Input: who sings the song Love Story}).
Fig.\ \ref{fig:instruction_detail} shows their average performances across all task categories. 
Even with trivial dataset tags, \ours outperforms the original GTR model, illustrating the effectiveness of instructions for diverse training.
As more information is provided in the instruction (from tag to simple and from simple to detail), we observe consistent improvements.




\subsection{Model Sizes and Instruction Finetuning}
\label{section:model_size}

\begin{figure}[!h]
    \centering
    \includegraphics[scale=0.32]{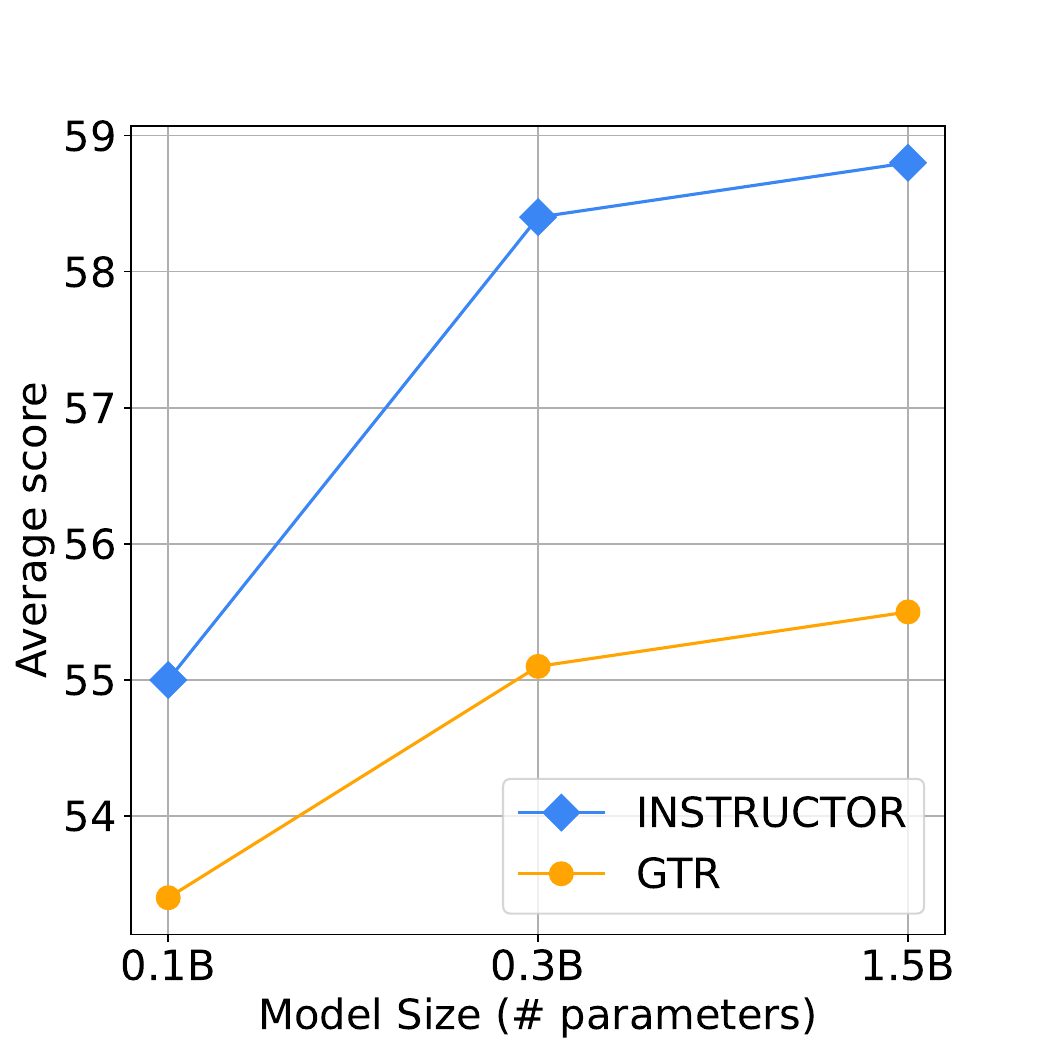} 
    \caption{Average performance comparisons with varying sizes of models.
    \ours benefits more from scaling up, perhaps because instructions require additional computations.
    }
    \label{fig:model_sizes}
\end{figure}

Fig.\ \ref{fig:model_sizes} studies the influence of model sizes.
Specifically, we use GTR-Base (0.1B), GTR-Large (0.3B), and GTR-XL (1.5B).
They are pretrained on the same corpus and differ only in the encoder size (the embedding sizes are the same).
We compare models of various sizes and report the average performance across all the categories.
As the encoder transformer model scales up, the performance continues to increase for both GTR and \ours.
Nonetheless, the improvement in \ours is more pronounced, perhaps because embeddings with instructions benefit from larger capacities.
This implies that large models are more generalizable to compute texts in various domains and task types, providing embeddings for general purposes.
Further scale-ups are left to future work.

\subsection{Instructions Mitigate Domain Shifts}
\label{section:domain_shift}
One advantage of instruction-based finetuning is that it improves models' ability to generalize to unseen domains and tasks.
To demonstrate this effectiveness, we found three unseen domains that \ours was not trained on: geography, biology, and civil comments.
As shown in Table \ref{tab:domain_specific}, \ours largely improves (above the average improvement) GTR-Large's performance on all three domains, indicating that instructions can help more when applying models to unseen or uncommon domains.

\begin{table}[!h]\centering
\small
\begin{tabular}{lcccc}\toprule
\textbf{Model} & Geography & Biology & Civil \ \\\midrule
GTR-Large & 53.4 &25.7 & 71.8 \\ 
\ours  & 64.2 & 31.3 & 77.2 \\
Relative gain (\%) & \textcolor{blue}{+20.2} & \textcolor{blue}{+21.8} & \textcolor{blue}{+7.5} \\
\bottomrule
\end{tabular} 
\caption{Results of GTR-Large and \ours on unseen domains: geography, biology and civil comments. Domain-specific datasets benefit particularly from instruction finetuning. 
More results can be found in Tables \ref{tab:detailed_results1} and \ref{tab:detailed_results2} in the appendix; they also show similar trends.   
}\label{tab:domain_specific}
\end{table}


\subsection{Qualititive Analysis}
\label{section:qual_analysis}

\begin{figure}[!h]
    \centering
    \includegraphics[scale=0.36]{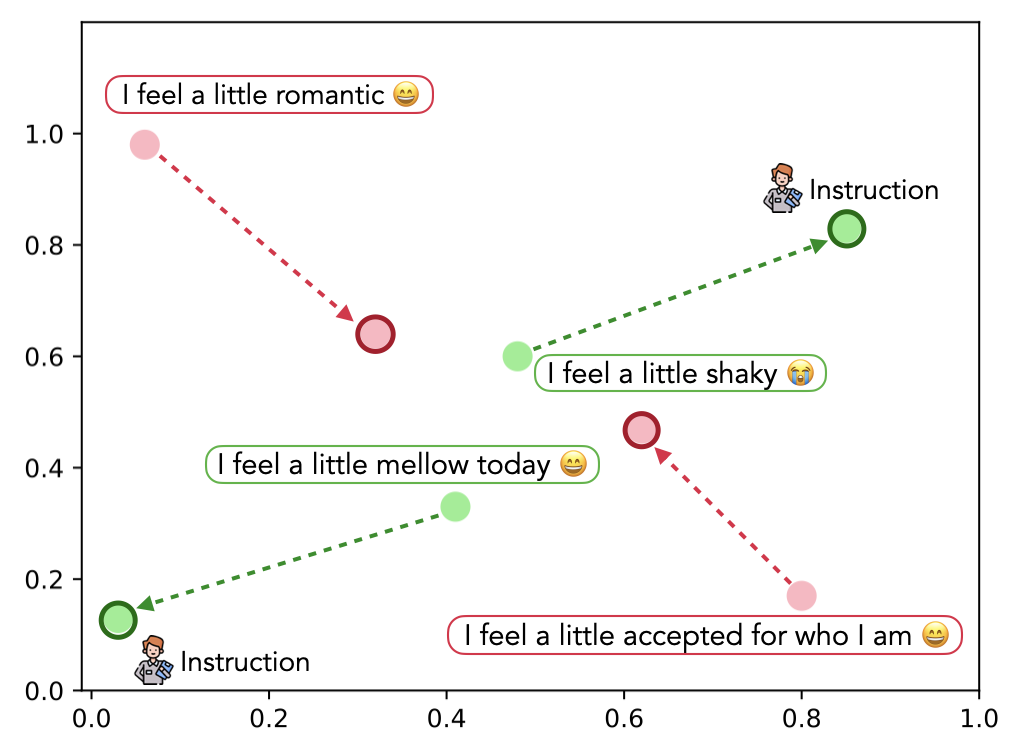} 
    \caption{
    Visualization of pair classification examples without (dot) and with instruction (dot with a solid border). The red dot pairs that have the same sentiment should be closer together, while the green dot pairs with different sentiment should be farther apart. When embedded with the instruction, the distance between the green dot pair becomes farther. 
    }
    \label{fig:qualitive}
\end{figure}

In this qualitative analysis, we use T-SNE \cite{tsne} to visualize two example of classification with and without instructions. 
The desired outcome is, for pairs with the same sentiment to be closer together, and pairs with different sentiment to be farther apart. 
As shown in Fig.~\ref{fig:qualitive}, without instructions, the green dot pairs (different sentiment) are closer together in the embedding space, while the red dot pairs (same sentiment) are farther apart. 
However, with instructions, our method (\ours) successfully encodes the red dot pairs into close embeddings and correctly classifies the pairs. 
The distance between the green dot pairs with different sentiment is also larger in the embedding space with instructions.

%% file: text/related.tex
\section{Related Work}
\paragraph{Text Embeddings} Text embeddings are useful in many applications such as information retrieval~\cite{thakur2021beir}, text similarity~\cite{gao-etal-2021-simcse}, prompt retrieval for in-context learning ~\cite{Selective_Annotation}, classification~\cite{reimers-gurevych-2019-sentence}, and beyond.
Much prior work develops different embedding models for different applications. For example, SBERT~\cite{reimers-gurevych-2019-sentence} and SimCSE~\cite{gao-etal-2021-simcse} are applied solely to text similarity and classification tasks, while DPR~\cite{karpukhin-etal-2020-dense} and Contriever~\cite{izacard2022unsupervised} focus on information retrieval.
Different from Sentence-T5 trained only on symmetric data or GTR trained only on asymmetric data, we combine both groups of datasets and build \dataset, which is then used to train \ours with instructions.
\citet{mteb} introduced the massive text embedding benchmark (MTEB), which can be used to evaluate embedding models on a variety of embedding tasks, spanning reranking, classification, information retrieval, bitext mining, pair classification, STS, and summarization.
Their benchmark shows that models performing well on one task may not perform well on other tasks.
The poor zero-shot transfer abilities of existing embedding models make it difficult to use them in applications where only few labeled data are available.
This motivates us to develop a single embedding model that is applicable to a variety of tasks and has better generalization to unseen tasks.
\citet{Wang2022TextEB} recently proposed E5, weakly-supervised contrastive pre-trained text embeddings, which achieve strong performance across various tasks on the MTEB benchmark, employing a larger embedding dimension compared to \ours.

\paragraph{Instruction Finetuning} 
Recent work demonstrated that instruction-finetuned language models could perform new tasks given a natural language instruction~\cite{mishra2022cross, Zhong2021AdaptingLM,min-etal-2022-metaicl,Sanh2022Multitask,Wei2022FinetunedLM, natural_instructions_v2,ouyang2022training}.
Nonetheless, instruction finetuning has yet to be studied in the context of broadly-applicable embeddings.
In this work, we explore finetuning embedding models to follow human instructions where the instruction specifies eventual use cases. 
Concurrent work demonstrated that instructions could facilitate information retrieval \cite{task_aware_retrieval}, which is related to our \ours design. 
They used instructions to build a task-aware retrieval system and conducted evaluations on the retrieval task; we build a general-purpose embedding model with instructions that can be applied to 8 tasks categories (Fig.\ \ref{fig:pipeline}), including retrieval, text similarity, clustering, and text evaluation.




%% file: text/future.tex
\section{Conclusion}
We introduced \ours, a single model that creates broadly-applicable text embeddings using natural language instructions.
We constructed \dataset, a collection of diverse datasets, to finetune \ours with instructions.
Our extensive experiments showed that \ours achieves state-of-the-art performance on text embedding benchmarks, as well as prompt retrieval for few-shot in-context learning.
We hope that researchers and practitioners will benefit from our embeddings or our datasets for tasks of their interest.

%% file: text/limitations.tex
\section{Limitations}


Although \ours significantly improves the baseline GTR performance, 
we were only able to use four negative examples during the model finetuning process due to computation constraints. However, negative examples have been shown to play an important role in contrastive learning~\cite{robinson2021contrastive}. We hope that future work will scale 
up the number of negatives used during finetuning and investigate various methods for mining hard negatives. Additionally, we do not have enough computation resources to apply multitask instruction finetuning to GTR-XXL (4.8B parameters), which is also an area for future exploration.

At the core of \ours is the instruction design. While our current unified instruction format has demonstrated effectiveness, future research can explore other instructional elements to further improve performance. For example, previous work~\cite{natural_instructions_v2} have shown that incorporating demonstration examples and explanations can be beneficial for instruction-finetuned language models. 


%% file: text/ack.tex
\section*{Acknowledgements}
We thank Akari Asai, Jack Lin, Minghan Li, and the ARK group at UW for their helpful feedback on this work.

%% file: text/appendix.tex
\begin{appendices}

\section{Training Setups}
\label{sec:training_settings}

\paragraph{Minibatch Sampling}
Training is performed on a combination of all training datasets in \dataset. Since the number of examples in each dataset is different in orders of magnitude, we downsample large ones.
Details for the downsampled numbers of examples on each dataset are shown in Table \ref{tab:data_statistics} in the appendix.
At each step, we first randomly select a dataset and then construct a minibatch \emph{only} using the examples from that dataset.
In this way, we ensure that in-batch negatives are sampled from the same dataset, thereby preventing the model from using task differences to predict the negative label.
We use the maximum batch size that fits the machine memory and run all our experiments on 40GB A100 GPUs.

\paragraph{Training}
We initialize \ours with the GTR-Large model~(\citealp{gtr21}, 335M parameters)\footnote{\url{https://huggingface.co/sentence-transformers/gtr-t5-large}.} and finetune it on \dataset using the AdamW optimizer with learning rate $2\times10^{-5}$ and warmup ratio 0.1.
We use a softmax temperature of 0.01 and finetune \ours for 20K steps. 

\paragraph{Baselines}
We use the official MTEB benchmark for comparisons, but here we highlight several strong baselines with the following two types.
The first class of baselines is embedding models specializing in information retrieval: \textbf{Contriever-MS}~\cite{izacard2022unsupervised}, \textbf{GTR}~\cite{gtr21}, and \textbf{coCondenser-MS}~\cite{gao2022unsupervised}.
They are all trained on open-domain QA datasets such as MS MARCO~\cite{bajaj2016ms}.
The second class of baselines focuses on  semantic textual similarity: \textbf{SimCSE}~\cite{gao-etal-2021-simcse}, \textbf{Sent-T5}~\cite{ni2022sentence}, and \textbf{SGPT-NLI}~\cite{muennighoff2022sgpt}. They are mainly trained on symmetric paraphrase datasets such as NLI~\cite{williams-etal-2018-broad} and the Quora question pairs.\footnote{\url{https://www.quora.com/q/quoradata/}.}
All of these baselines are based on pretrained language models, achieving strong performance on the MTEB leaderboard.
In particular, Sent-T5-XXL and GTR-XXL (both with 4.8B parameters) achieve the first and second best average performances.

\section{Embedding Evaluations}
\label{sec:eval_settings}
Here we provide a high-level summary of the evaluation tasks (Table~\ref{tab:instruct}).
Following MTEB \cite{mteb}, Billboard \cite{billboard}, and prompt retrieval \cite{Selective_Annotation}, we split 70 evaluation datasets into 9 categories by task objectives.
Out of the 70 evaluation tasks, 66 are unseen during training (See \autoref{tab:data_statistics} for datasets included during training), 
Table \ref{tab:instruct} for examples and instructions for the evaluation datasets.

\subsection{Massive Text Embedding Benchmark}
\label{section:mteb}
MTEB~\cite{mteb} is a comprehensive embedding evaluation benchmark that aims to provide a holistic view of current embedding models' performance and to discover universal text embeddings applicable to a wide range of tasks.
It combines several conventional benchmarks (e.g., BEIR, \citealp{thakur2021beir}, and STS, \citealp{cer2017semeval}) and spans a wide range of domain-specific datasets, including science, biology, and medicine.
Following \citet{mteb}, we also report the average performance over 56 datasets.
For each task family, we briefly describe the task objective, evaluation metric, and how embeddings are used. 

\paragraph{Retrieval}
Given a query $q$ and a corpus $D=\{p_1, p_2...p_n\}$, retrieval aims to find the most relevant documents $p_i$ in $D$ for query $q$. 
The embedding model is used to embed $q$ and $p_1...p_n$ into fixed-sized vectors, and then the similarity between $q$ and $p_i$ is measured by their embedding cosine similarity. 
There are 14 diverse datasets (e.g., Natural Questions, Scifact, and NFCorpus) together with the community question-answering (CQA) benchmark \cite{hoogeveen2015cqadupstack}.
We use NDCG@10 (Normalized Discounted cumulative gain at rank position 10) to measure the performance.

\paragraph{Reranking}
Reranking ranks a list of documents based on their relevance to a query. 
Given a query $q$ and a list of documents $D=\{p_1, p_2...p_n\}$, the embedding model computes embeddings of both the query and documents, which are then used to rank the documents based on their cosine similarities.
We use MAP (mean average precision), a standard metric in reranking, to measure performance.

\paragraph{Clustering}  
The goal of clustering is to group similar documents into meaningful clusters. 
Given a set of documents, the encoder maps each document into an embedding.
The k-means clustering algorithm is then used to partition the embedded documents into clusters.
The clustering performance is measured by the v-measure that is independent of the permutations of clustering labels \cite{rosenberg-hirschberg-2007-v}. 

\paragraph{Pair Classification}
Pair classification tasks aim to predict a binary label for a pair of texts.
An example of this task is paraphrase identification, where the goal is to predict whether two sentences are paraphrases of each other.
Given a sentence pair $(t_1, t_2)$, the embedding model encodes $t_1$ and $t_2$ separately.
The cosine similarity between the two embeddings is then used to predict the label.
The average precision score is measured for evaluation.

\paragraph{Classification}
Classification is a popular way to evaluate the quality of embeddings~\cite{conneau-kiela-2018-senteval}.
For each example in the classification dataset, the embedding of the input text is used as features to a classifier.
The classifier is trained on the training data while sentence embedings are kept frozen.
We report the classification accuracy on the test set as the evaluation metric.

\paragraph{STS}
Semantic textual similarity (STS) tasks evaluate the similarity between two sentences. Given a sentence pair $(t_1, t_2)$, the embedding model maps $t_1$ and $t_2$ into embeddings separately, and then the similarity between $t_1$ and $t_2$ is measured by their embedding cosine similarity.
The evaluation metric is Spearman's rank correlation, which measures the correlation between the similarity scores and human judgements.

\paragraph{Summarization}
Automatic summarization evaluation aims to evaluate the quality of a machine-generated summary given a reference summary.
While human evaluations are considered more accurate, automatic evaluations allow for fast, inexpensive development cycles \cite{genie}.
Given a reference summary $r$ and a machine-generated summary $t$, the embedding model maps them into embeddings separately, and we compute the cosine similarity between $r$ and $t$. 
Spearman's rank correlation is reported between human judgements and automatic scores.


\begin{table}
\small
\begin{tabular}{lm{5cm}}
\toprule
& \hspace{2cm}Examples \\
\midrule
Text type & question, query, answer, summary, sentence, review, post, comment, statement, paragraph, passage, document \\
\midrule
Text objective & classify the sentence as positive or negative, retrieve a duplicate sentence, retrieve the supporting document\\
\midrule
Domain & wikipedia, news, medicine, biology, reddit, stackoverflow, science, quora,  coronavirus, math, physics\\
\bottomrule
\end{tabular}
\caption{Examples of text types, objectives, and domains.
}
\label{tab:instruction_examples}
\end{table}
\subsection{Prompt Retrieval}
\label{section:in_context}
Large language models have demonstrated the ability of in-context learning, where the model can perform downstream tasks by conditioning generation on a few task demonstrations \cite{prompt_survey}. \citet{Selective_Annotation} introduce the prompt retrieval task, where the goal is to retrieve a few in-context learning (i.e., demonstration) examples from annotated examples given a test instance.
The embedding model is used to encode all annotated examples and to find the few most similar examples to the test instance based on the cosine similarity.
Following \citet{Selective_Annotation}, we use the retrieved examples for in-context learning on GPT-J \cite{gpt-j} over 11 diverse downstream tasks (e.g., classification, multiple choice, and text-to-SQL) 
that are not included in \dataset (thus zero-shot settings).
We compare different embedding methods by measuring the average performance on these downstream tasks.

\input{table/data_statistics.tex}

\subsection{Automatic Evaluation for Generation}
\label{section:text_evaluation}
Similar to summarization evaluation in MTEB, we use the Billboard benchmark \cite{billboard} to apply \ours to automatic evaluations for three additional text generation tasks: MSCOCO image captioning \cite{mscoco,thumb}, CNN/DailyMail news summarization~\cite{fabbri2021summeval}, and WMT21 Chinese-to-English translation \cite{barrault-etal-2020-findings,freitag2021experts}.
Following \citet{billboard}, we measure the cosine similarity between the generated text and each reference text and take the maximum similarity score over all references available \cite{bertscore}. 
We evaluate all embedding models by the Pearson correlation with the human judgments, again following \citet{billboard}.
We then report the average correlation scores over the three datasets.
Note that we do not use the English-to-German dataset in Billboard because our models are trained only on English data.

\input{table/prompt_variants.tex}
\section{Full instructions}  
We list all instructions for each dataset in \dataset in Table \ref{tab:full_instructions1} and Table \ref{tab:full_instructions2}
\label{appendix:full instruction}
\input{table/full_instruction.tex}

\section{Full Results}
We provide the detailed evaluation scores in MTEB, Billboard and prompt retrieval benchmarks in Table \ref{tab:detailed_results1} \& \ref{tab:detailed_results2}.

\input{table/detailed_score_1.tex}

\input{table/detailed_score_2.tex}

\end{appendices}

%% file: table/data_statistics.tex
\begin{table*}[hbt!]
\small
\centering
\begin{tabular}{lcl}
\toprule
Dataset & Symmetric/Asymmetric &Number \\
\midrule
gooaq\_pairs~\cite{gooaq2021} & Asymmetric&25,000 \\
yahoo\_answers\_title\_answer~\cite{zhang2015character}& Asymmetric&25,000 \\
stackexchange~\cite{silva2018duplicate}& Symmetric&25,000 \\
eli5\_question\_answer~\cite{eli5_lfqa}& Asymmetric&25,000 \\
squad\_pairs~\cite{2016arXiv160605250R}& Asymmetric&25,000 \\
NQ$^*$~\cite{kwiatkowski2019natural}& Asymmetric&50,000 \\
amazon-qa~\cite{gupta2019amazonqa}& Asymmetric&100,000 \\
WikiAnswers~\cite{Fader14}& Symmetric&25,000 \\
agnews~\cite{zhang2015character}& Asymmetric&45,000 \\
AllNLI~\cite{bowman2015large}& Symmetric&50,000 \\
npr~\cite{shuyang}& Asymmetric&25,000 \\
specter\_train\_triples~\cite{specter2020cohan}& Symmetric&50,000 \\
ccnews\_title\_text~\cite{Hamborg2017}& Asymmetric&25,000 \\
triviaqa~\cite{2017arXivtriviaqa}& Asymmetric&50,000 \\
zero\_shot\_re~\cite{levy2017zero}& Asymmetric&15,000 \\
flickr30k\_captions~\cite{young2014image}& Symmetric&25,000 \\
xsum~\cite{Narayan2018DontGM}& Asymmetric&10,000 \\
code\_search~\cite{husain2019codesearchnet}& Asymmetric&15,000 \\
msmarco$^*$~\cite{bajaj2016ms}& Asymmetric&175,000 \\
hotpotqa$^*$~\cite{yang2018hotpotqa}& Asymmetric&40,000 \\
fever$^*$ ~\cite{thorne2018fever}& Asymmetric&75,000 \\
amazon\_review\_2018 ~\cite{he2016ups}& Asymmetric&100,000 \\
S2ORC\_title\_abstract~\cite{lo-wang-2020-s2orc}& Asymmetric&100,000 \\
PAQ\_pairs~\cite{lewis2021paq}& Asymmetric&25,000 \\
wow~\cite{dinan2018wizard}&Asymmetric &30,000 \\
trex~\cite{elsahar2018t}& Asymmetric&30,000 \\
pubmed~\cite{sen2008collective}&Asymmetric &30,000 \\
medmcqa~\cite{pmlr-v174-pal22a}& Asymmetric&30,000 \\
wikihow~\cite{koupaee2018wikihow}& Asymmetric&5,000 \\
simple\_wiki~\cite{coster2011simple}& Asymmetric&5,000 \\
Super-NI (300 datasets)~\cite{natural_instructions_v2}& Symmetric&180,000 \\
\bottomrule
\end{tabular}
\caption{Number of training instances in each dataset. The dataset with * indicates that its test-split is included in the evaluation.
}
\label{tab:data_statistics}
\end{table*}

%% file: table/prompt_variants.tex
\begin{table*}[hbt!]
\small
\begin{tabular}{cm{14.5cm}}
\toprule
Dataset &  \hspace{5.5cm}Instruction \\
\midrule
& \makecell[l]{\textbf{\textit{Instruction 1:}} Represent the Amazon comment for classifying the sentence as positive or negative:}\\
Amazon & \makecell[l]{\textbf{\textit{Instruction 2:}} Represent the Amazon review comment for classifying the emotion as positive or negative:}\\
Polarity & \makecell[l]{\textbf{\textit{Instruction 3:}} Represent the Amazon sentence for classifying its sentiment as positive or negative:}\\
Classification & \makecell[l]{\textbf{\textit{Instruction 4:}} Represent an Amazon post for classifying its sentiment as positive or negative:}\\
& \makecell[l]{\textbf{\textit{Instruction 5:}} Represent the Amazon review for classifying the review sentiment as negative or positive:}\\
\midrule
& \makecell[l]{\textbf{\textit{Query instruction 1:}} Represent the finance query for retrieving supporting documents: \\ \textbf{\textit{Doc instruction 1}}: Represent the finance document for retrieval:}\\
& \makecell[l]{\textbf{\textit{Query instruction 2:}} Represent the financial question for retrieving supporting documents:\\ \textbf{\textit{Doc instruction 2}}: Represent the financial document for retrieval:}\\
FIQA2018 & \makecell[l]{\textbf{\textit{Query instruction 3:}} Represent the finance query for retrieving related documents: \\ \textbf{\textit{Doc instruction 3}}: Represent the finance document for retrieval:}\\
& \makecell[l]{\textbf{\textit{Query instruction 4:}} Represent a finance query for retrieving relevant documents: \\ \textbf{\textit{Doc instruction 4}}: Represent the financial document for retrieval:}\\
& \makecell[l]{\textbf{\textit{Query instruction 5:}} Represent the finance query for retrieving supporting passages: \\ \textbf{\textit{Doc instruction 5}}: Represent the finance passage for retrieval:}\\

\bottomrule
\end{tabular}
\caption{Example paraphrased instructions for AmazonPolarityClassification and FIQA2018. They follow the unified template (\S\ref{section:unified_instruction_format}) with the same information and only differ in wording choices.
}
\label{tab:prompt_variants}
\end{table*}



%% file: table/full_instruction.tex
\begin{table*}[hbt!]
\small
\begin{tabular}{cm{11cm}}
\toprule
Dataset &  \hspace{5.5cm}Instruction \\
\midrule
 MSMARCO & \makecell[l]{\textbf{\textit{Query instruction:}} Represent the [domain] question for retrieving evidence documents: \\ \textbf{\textit{Doc instruction}}: Represent the {domain} document for retrieval:}
 \\
 \midrule
 gooaq\_pairs & \makecell[l]{\textbf{\textit{Query instruction:}} Represent the Google question for retrieving answers:\\ \textbf{\textit{Doc instruction}}: Represent the Google answer for retrieval:}\\
 \midrule
  yahoo\_answers\_title\_answer & \makecell[l]{\textbf{\textit{Query instruction:}} Represent the Yahoo question for retrieving answers: \\
 \textbf{ \textit{Doc instruction:}} Represent the Yahoo answer for retrieval:}\\
  \midrule
  eli5\_question\_answer & \makecell[l]{\textbf{\textit{Query instruction:}} Represent the ELI5 question for retrieving answers: \\ \textbf{\textit{Doc instruction:}} Represent the ELI5 answer for retrieval:}\\
  \midrule
  squad\_pairs &  \makecell[l]{\textbf{\textit{Query instruction:}} Represent the Squad question for retrieving evidence documents: \\ 
 \textbf{ \textit{Doc instruction:}} Represent the Squad document for retrieval:}\\
  \midrule
  Natural Question & \makecell[l]{\textbf{\textit{Query instruction:}} Represent the Wikipedia question for retrieving supporting documents: \\
 \textbf{ \textit{Doc instruction:}} Represent the Wikipedia document for retrieval:} \\
  \midrule
  amazon-qa & \makecell[l]{\textbf{\textit{Query instruction:}} Represent the Amazon question for retrieving answers: \\ \textbf{\textit{Doc instruction:}} Represent the Amazon answer for retrieval:}\\
  \midrule
  agnews & \makecell[l]{\textbf{\textit{Query instruction:}} Represent the news title for retrieving relevant articles: \\
  \textbf{\textit{Doc instruction:} }Represent the news article for retrieval:}\\
  \midrule
  npr & \makecell[l]{\textbf{\textit{Query instruction:}} Represent the news title for retrieving relevant articles: \\ \textbf{\textit{Doc instruction:}} Represent the news article for retrieval:}\\
  \midrule
  ccnews\_title\_text & \makecell[l]{\textbf{\textit{Query instruction:}} Represent the news title for retrieving relevant articles: \\ \textbf{\textit{Doc instruction:}} Represent the news article for retrieval:}\\
  \midrule
  triviaqa & \makecell[l]{\textbf{\textit{Query instruction:}} Represent the question for retrieving evidence documents: \\ \textbf{\textit{Doc instruction}}: Represent the evidence document for retrieval:}\\
  \midrule
  zero\_shot\_re & \makecell[l]{\textbf{\textit{Query instruction:}} Represent the Wikipedia question for retrieving evidence documents: \\ \textbf{\textit{Doc instruction:}} Represent the Wikipedia document for retrieval:}\\
  \midrule
  xsum & \makecell[l]{\textbf{\textit{Query instruction:}} Represent the news title for retrieving relevant articles: \\ \textbf{ \textit{Doc instruction:}} Represent the news article for retrieval:}\\
  \midrule
  code\_search & \makecell[l]{\textbf{\textit{Query instruction:}} Represent the comment for retrieving corresponding codes: \\ \textbf{\textit{Doc instruction:}} Represent the code for retrieval:}\\
  \midrule
  hotpotqa & \makecell[l]{\textbf{\textit{Query instruction:}} Represent the Wikipedia question for retrieving supporting documents: \\ \textbf{\textit{Doc instruction:}} Represent the Wikipedia document for retrieval:}\\
  \midrule
  fever & \makecell[l]{\textbf{\textit{Query instruction:}} Represent the fact for retrieving supporting evidence:\\ \textbf{\textbf{\textit{Doc instruction:}}} Represent the evidence for retrieval:}\\
  \midrule
  amazon\_review\_2018 & \makecell[l]{\textbf{\textit{Query instruction:}} Represent the Amazon title for retrieving relevant reviews: \\ \textbf{\textit{Doc instruction:}} Represent the Amazon review for retrieval:}\\
  \midrule
  S2ORC\_title\_abstract & \textbf{\textit{Query instruction:}} Represent the Scientific title for retrieving relevant abstracts:, 
  \textbf{\textit{Doc instruction}:} Represent the Scientific abstract for retrieval:\\
  \midrule
  PAQ\_pairs &  \textbf{\textit{Query instruction:}} Represent the question for retrieving evidence documents:,  \textbf{\textit{Doc instruction}:} Represent the evidence document for retrieval:\\
  \midrule
  wow & \textbf{\textit{Query instruction:}} Represent the Wikipedia question for retrieving supporting documents:,\textbf{ \textit{Doc instruction}:} Represent the Wikipedia document for retrieval:\\
  \midrule
  trex & \textbf{\textit{Query instruction:}} Represent the Wikipedia question for retrieving supporting documents:, \textbf{\textit{Doc instruction}:} Represent the Wikipedia document for retrieval:\\
  \midrule
  pubmed & \textbf{\textit{Query instruction:}} Represent the Medicine sentence for retrieving relevant documents:, \textbf{\textit{Doc instruction}:} Represent the Medicine document for retrieval:\\
  \midrule
  medmcqa & \textbf{\textit{Query instruction:}} Represent the Medicine question for retrieving supporting answers:, \textbf{\textit{Doc instruction:}} Represent the Medicine answer for retrieval:\\
  \midrule
  wikihow & \textbf{\textit{Query instruction:}} Represent the Wikipedia summary for retrieving relevant passages:, \textbf{\textit{Doc instruction}:} Represent the Wikipedia passage for retrieval:\\
  \midrule
  simple\_wiki & \textbf{\textit{Query instruction:}} Represent the Wikipedia sentence for retrieving simplified sentences:, \textbf{\textit{Doc instruction}:} Represent the Wikipedia sentence for retrieval:\\
\bottomrule
\end{tabular}
\caption{Instructions of asymmetric training dataset. We use Kmeans clustering to put MSMARCO examples into 30 groups, and label the domain for each group.}
\label{tab:full_instructions1}
\end{table*}

\begin{table*}[hbt!]
\small
\begin{tabular}{cm{12cm}}
\toprule
Dataset &  \hspace{5.5cm}Instruction \\
  \midrule
  stackexchange &  \textbf{\textit{Instruction:}} Represent the StackExchange question for retrieving duplicate questions:\\
  \midrule
  WikiAnswers & \makecell[l]{\textbf{\textit{Instruction:}} Represent the Wikipedia question for retrieving duplicate questions:}\\
  \midrule
  AllNLI & \makecell[l]{\textbf{\textit{Instruction:}} Represent the sentence for retrieving duplicate sentences:}\\
  \midrule
  specter\_train\_triples &  \makecell[l]{\textbf{\textit{Instruction:}} Represent the scientific title for retrieving duplicate titles:}\\
  \midrule
  flickr30k\_captions & \makecell[l]{\textbf{\textit{Instruction:}} Represent the caption for retrieving duplicate captions:}\\
  \midrule
  super-NI & \textbf{\textit{Instruction:}} Represent the example for the following task: [dataset definition]:\\

\bottomrule
\end{tabular}
\caption{Instructions of symmetric training dataset. We use the task definitions of Super-NaturalInstructions as the task objective.
}
\label{tab:full_instructions2}
\end{table*}


%% file: table/detailed_score_1.tex
\begin{table*}[h]
\small
 \addtolength{\tabcolsep}{-1pt} 
\centering
\begin{tabular}{m{2.3cm}m{5cm}cccc}
\toprule
Category & Dataset & GTR &  \ours & GTR & \ours  \\
         &         & 335M &  335M & 1.5B &1.5B  \\
\midrule
&SciFact                                     & 63.8   & 64.3 & 64.2 & 64.6  \\
&NFcorpus                                    & 32.4 & 34.1 & 33.3 & 36.0  \\
&ArguAna                                     & 52.1 & 57.1 & 52.8 & 55.7  \\
&CQADupstackWebmastersRetrieval              & 35.7 & 46.4 & 36.5 & 45.1  \\
&CQADupstackEnglishRetrieval                 & 46.8 & 50.8 & 46.5 & 49.3  \\
&CQADupstackGamingRetrieval                  & 56.3 & 63.1 & 55.8 & 63.3  \\
&CQADupstackGisRetrieval                     & 33.7 & 39.5 & 34.6 & 40.6  \\
&CQADupstackAndroidRetrieval                 & 46.1 & 55.9 & 44.9 & 55.0  \\
&CQADupstackTexRetrieval                     & 25.1 & 30.0 & 26.1 & 29.1  \\
&CQADupstackUnixRetrieval                    & 35.3 & 44.7 & 36.6 & 42.5  \\
&CQADupstackMathematicaRetrieval             & 24.8 & 30.7 & 27.4 & 30.8  \\
&CQADupstackStatsRetrieval                   & 30.4 & 34.6 & 30.1 & 35.7  \\
Retrieval&CQADupstackPhysicsRetrieval        & 38.5 & 47.8 & 39.7 & 45.3  \\
&CQADupstackProgrammersRetrieval            
& 38.5 & 47.5 & 39.6 & 44.9  \\
&CQADupstackWordpressRetrieval               & 28.2 & 34.9 & 30.4 & 35.5  \\
&ClimateFEVER                                & 26.9 & 27.8 & 27.0 & 26.5  \\
&FEVER                                       & 72.7 & 72.7 & 72.2 & 70.0  \\
&FiQA2018                                    & 42.8 & 45.5 & 44.2 & 47.0  \\
&HotpotQA                                    & 57.9 & 55.2 & 58.9 & 55.9  \\
&MSMARCO                                     & 42.7 & 39.7 & 43.5 & 41.6  \\
&NQ                                          & 55.1 & 50.1 & 56.2 & 57.3  \\
&QuoraRetrieval                              & 88.5 & 88.4 & 88.9 & 88.9 \\
&SCIDOCS                                     & 15.5 & 18.6 & 15.7 & 17.4  \\
&DBPedia                                     & 39.6 & 36.7 & 39.7 & 40.2  \\
&TRECCOVID                                   & 56.7 & 58.1 & 60.1 & 71.4  \\
&Touche2020                                  & 28.3 & 21.6 & 25.3 & 23.4  \\
\bottomrule
\end{tabular}
\caption{All Retrieval results in MTEB benchmark.
}
\label{tab:detailed_results1}
\end{table*}

%% file: table/detailed_score_2.tex
\begin{table*}[h]
\small
\begin{tabular}{m{2.3cm}m{5cm}cccc}
\toprule
Category & Dataset & GTR & \ours & GTR & \ours \\
         &         & 335M & 335M & 1.5B & 1.5B \\
\midrule
Summarization&SummEval                       & 29.5 & 31.8 & 30.2 & 32.0  \\
\midrule
&AskUbuntuDupQuestions                       & 61.6  & 64.3 & 63.1 & 65.4  \\
Reranking&StackOverflowDupQuestions          & 51.6 & 52.2 & 52.8 & 52.5  \\
&SciDocsRR                                   & 76.4 & 82.0 & 76.5 & 79.5  \\
&MindSmallReranking                          & 31.8 & 31.7 & 31.5 & 31.8  \\
\midrule
&BiorxivClusteringS2S                        & 25.7  & 31.3 & 26.1 & 30.6  \\
&MedrxivClusteringS2S                        & 27.4 & 32.0 & 26.7 & 30.8  \\
&TwentyNewsgroupsClustering                  & 51.6 & 54.1 & 51.2 & 53.3  \\
&ArxivClusteringP2P                          & 37.5 & 43.2 & 37.9 & 42.5  \\
&ArxivClusteringS2S                          & 30.6 & 32.6 & 30.5 & 32.2  \\
Clustering&BiorxivClusteringP2P              & 29.6 & 37.6 & 30.5 & 37.5  \\
&MedrxivClusteringP2P                        & 28.7 & 34.2 & 28.7 & 33.2  \\
&RedditClustering                            & 61.7 & 63.7 & 61.3 & 63.4  \\
&RedditClusteringP2P                         & 61.7 & 64.6 & 61.1 & 65.1  \\
&StackExchangeClustering                     & 69.9 & 68.8 & 70.0 & 68.4  \\
&StackExchangeClusteringP2P                  & 33.2 & 36.1 & 32.7 & 35.1  \\
\midrule
&SprintDuplicateQuestions                    & 95.1 & 93.1 & 95.5 & 94.9  \\
Pair Classification&TwitterSemEval2015       & 76.0 & 77.4 & 77.8 & 78.0  \\
&TwitterURLCorpus                            & 84.9 & 87.2 & 85.1 & 86.9 \\
\midrule
&STS12                                       & 70.3 & 76.3 & 69.1 & 75.3  \\
&STS13                                       & 82.2 & 88.2 & 81.8 & 87.4  \\
&STS14                                       & 77.2 & 81.9 & 77.1 & 81.9  \\
&STS15                                       & 86.3 & 89.0 & 86.0 & 88.9  \\
STS&STS16                                    & 81.9 & 85.5 & 82.2 & 85.4  \\
&STS17                                       & 83.9 & 90.3 & 84.9 & 90.5  \\
&STS22                                       & 64.3 & 67.4 & 66.6 & 68.6  \\
&BIOSSES                                     & 84.9 & 84.4 & 78.9 & 84.2  \\
&SICK-R                                      & 73.4 & 81.3 & 73.6 & 81.7  \\
&STSBenchmark                                & 77.6 & 86.9 & 77.7 & 86.6  \\
\midrule
&Banking77Classification                     & 81.2 & 78.5 & 82.2 & 82.7  \\
&TweetSentimentExtractionClassification      & 54.1 & 64.1 & 54.8 & 61.7  \\
&AmazonReviewsClassification                 & 37.2 & 47.9 & 38.2 & 43.0  \\
&EmotionClassification                       & 46.3 & 52.7 & 45.5 & 53.2  \\
&AmazonCounterfactualClassification          & 70.0 & 88.1 & 68.6 & 85.1  \\
Classification&ImdbClassification            & 70.9 & 88.3 & 68.2 & 80.1  \\
&MassiveIntentClassification                 & 70.1 & 68.9 & 70.2 & 71.5  \\
&MassiveScenarioClassification               & 75.5 & 73.4 & 75.9 & 76.5  \\
&MTOPIntentClassification                    & 63.9 & 68.0 & 65.9 & 72.3  \\
&MTOPDomainClassification                    & 94.0 & 93.9 & 93.6 & 95.1  \\
&AmazonPolarityClassification                & 73.9 & 91.5 & 74.6 & 86.5  \\
&ToxicConversationsClassification            & 68.7 & 71.1 & 67.6 & 70.3  \\
\midrule
&RTE                                         & 56.1 & 58.8 & 56.8 & 59.3  \\
&SST-5                                       & 52.4 & 53.8 & 53.2 & 60.1  \\
&coda19\_title\_generation                   & 21.2 & 23.6 & 21.4 & 27.8  \\
&multirc\_answerability                      & 62.5 & 63.6 & 63.7 & 72.6  \\
&MRPC                                        & 60.3 & 65.4 & 60.8 & 72.9  \\
Prompt Retrieval&HellaSwag                   & 61.6 & 62.8 & 63.4 & 72.4  \\
&Amazon                                      & 36.0 & 38.0 & 36.0 & 48.0  \\
&Dbpedia\_14                                 & 91.7 & 93.0 & 91.7 & 94.0  \\
&GeoQuery                                   & 53.4 & 64.2 & 53.5 & 63.2  \\
&Multi-Woz                                   & 90.8 & 94.4 & 91.0 & 95.2  \\
&CivilComments                               & 71.8 & 77.2 & 72.6 & 88.3  \\
\midrule
&mscoco                                      & 32.3 & 41.6 & 33.2 & 39.7  \\
Billboard&cnn summary                        & 25.8 & 30.3 & 26.1 & 31.9  \\
&machine translation                         & 35.4 & 38.9 & 36.6 & 30.6  \\

\bottomrule
\end{tabular}
\caption{All Prompt retrieval, Billboard, and MTEB English results, cont.
}
\label{tab:detailed_results2}
\end{table*}